\documentclass[lettersize,journal]{IEEEtran}
\usepackage{amsmath,amsfonts}
\usepackage{algorithmic}
\usepackage{algorithm}
\usepackage{array}
\usepackage[caption=false,font=normalsize,labelfont=sf,textfont=sf]{subfig}
\usepackage{textcomp}
\usepackage{stfloats}
\usepackage{url}
\usepackage{verbatim}
\usepackage{graphicx}
\usepackage{cite}
\usepackage{booktabs}
\usepackage{makecell}
\usepackage{multirow}
\usepackage{color}
\begin{document}

\title{Time-Division Multiplexing Actuation in Tendon-Driven Arms: Lightweight Design and Fault Tolerance}

\author{Shoujie Li$^{*}$, Changqing Guo$^{*}$,~\IEEEmembership{Student Member,~IEEE}, Jianle Xu$^{*}$, Hong Luo, Xueqian Wang,~\IEEEmembership{Member,~IEEE}, Wenbo Ding,~\IEEEmembership{Member,~IEEE},
Bin Liang,~\IEEEmembership{Senior Member,~IEEE}
\thanks{
This work was supported by National Key R\&D Program of China grant (2024YFB3816000),
Guangdong Innovative and Entrepreneurial Research Team Program (2021ZT09L197), National Natural Science Foundation of China (62003188), Tsinghua Shenzhen International Graduate School-Shenzhen Pengrui Young Faculty Program of Shenzhen Pengrui Foundation (No. SZPR2023005) and Meituan. 
\textit{*Shoujie Li, Changqing Guo, and Jianle Xu contributed equally to this work.} (Corresponding author: Xueqian Wang \&  Wenbo Ding,  wang.xq@sz.tsinghua.edu.cn \&  ding.wenbo@sz.tsinghua.edu.cn)}

\thanks{Shoujie Li, Changqing Guo, Jianle Xu, Hong Lu, Xueqian Wang, Wenbo Ding are with Shenzhen International Graduate School, Tsinghua University, Shenzhen 518055, China.}

\thanks{Shoujie Li is also with the School of Mechanical and Aerospace Engineering, Nanyang Technological University, Singapore 639956, Singapore.}
\thanks{Bin Liang is with the Department of Automation, Tsinghua University, Beijing 100084, China. }}



\maketitle

\begin{abstract}
Robotic manipulators for aerospace applications require a delicate balance between lightweight construction and fault-tolerant operation to satisfy strict weight limitations and ensure reliability in remote, hazardous environments. This paper presents Time-Division Multiplexing Actuation (TDMA), a practical approach for tendon-driven robots that significantly reduces actuator count while preserving high torque output and intrinsic fault tolerance. 
The key hardware employs a vertically-stacked rotational selection structure that integrates self-rotating TDM motors for rapid configuration, electromagnetic clutches enabling sub-0.1 second engagement, a worm gear reducer for enhanced load capacity and self-locking capability, and a dual-encoder system for precise, long-term positioning. Leveraging TDMA, the proposed MuxArm achieves a self-weight of 2.17 kg, supports an actuator driving capacity of 10 kg, and maintains end-effector accuracy up to 1\% of its length, even under partial servo failure. Additionally, an actuation space trajectory planning algorithm is developed, enabling fault-tolerant control and reducing tendon load by up to 50\% compared to conventional methods.  Comprehensive experiments demonstrate MuxArm’s robust performance in diverse settings, including free-space, cluttered, and confined environments. 
\end{abstract}

\begin{IEEEkeywords}
Tendon-Driven,  Time-Division, Lightweight, Fault-Tolerant
\end{IEEEkeywords}

\section{Introduction}
\IEEEPARstart{W}{ith} the continuous advancement of aerospace technology, flexible robotic arms have shown great potential and value in space applications\cite{zhang2023progress}. These robotic arms not only perform various tasks in complex space environments but also exhibit strong adaptability and operational flexibility. When designing space robotic arms, weight and stability are two critical factors \cite{11032106}. A lightweight design not only helps reduce launch costs but also enhances the control performance and stability of the robotic arm in microgravity environments. Moreover, the complex radiation and electromagnetic interference in space pose significant challenges to the stability of the driving motors\cite{flores2014review}. Therefore, designing a lightweight robotic arm with high payload capacity and fault tolerance is of great significance for advancing aerospace endeavours.

\begin{figure}[t]
    \centering
    \includegraphics[width=0.95\linewidth]{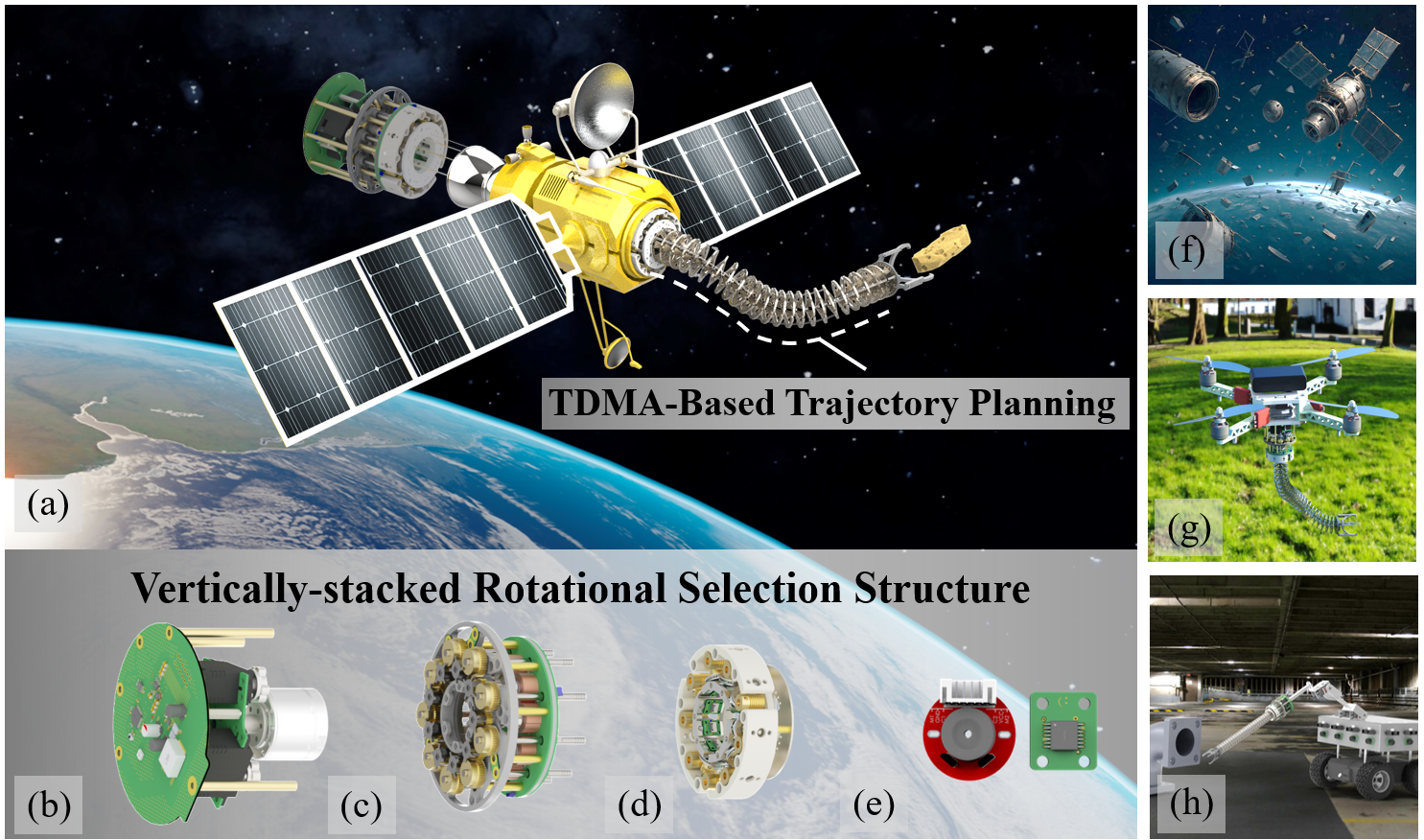}
    \caption{ Overview of the tendon-driven robotic arm based Time-Division Multiplexing Actuation (TDMA). (a) Concept illustration of the robotic arm mounted on a satellite for space exploration with TDMA-based trajectory planning. (b–e) Key hardware modules enabling TDMA through vertically-stacked rotational selection structure: (b) Rotational TDM motors; (c) Rapid-switching clutches; (d) High-torque winding units; (e) Dual-encoder sensing systems. (f–h) Potential application scenarios: (f) Space debris cleanup; (g) Unmanned Aerial Vehicle (UAV) robotic arm grasping; (h) Radiation environment hazard management for autonomous vehicles.}
    \label{fig:1}
\end{figure}

To address lightweight and fault-tolerant design challenges, researchers have explored structural solutions (origami-inspired guides~\cite{19}, modular backbones~\cite{17}), routing optimizations (helical paths~\cite{13}, constant-length routing~\cite{20}), and advanced actuators (magnetorheological clutches~\cite{12}, base-mounted systems~\cite{4}). Redundant configurations~\cite{25}, control algorithms~\cite{7}, and reconfigurable designs~\cite{5} enhance reliability but increase complexity and mass. Recent time-shared actuation~\cite{23} reduces actuator count yet suffers from frequent switching and auxiliary dependencies. These limitations motivate a mechanism integrating lightweight design, high torque capacity, and intrinsic fault tolerance.

To solve these problems, this paper introduces a vertically-stacked rotational Time-Division Multiplexing Actuation (TDMA), inspired by time-division communication principles, as shown in Fig.~\ref{fig:1}. TDMA employs a scheduler to dynamically multiplex multiple motors onto tendons, enabling multi-DOF control with reduced actuator count. This mechanism inherently improves fault tolerance by redistributing torque coupling among remaining servos during actuator failure. By reducing actuator count, TDMA enables integration of higher-capacity motors while the base-mounted configuration minimizes arm inertia, facilitating compact, lightweight, and cost-effective design for space exploration and orbital maintenance. The innovation lies in a novel vertically-stacked rotational selection mechanism that enables time-multiplexed actuation through co-designed hardware and algorithms. The primary contributions of this work are:

\begin{itemize}

\item \textbf{Selection mechanism.} A vertically-stacked rotational selection structure achieves efficient actuation, enabling control of more DOFs with fewer motors through rapid radial tendon selection via axial motor rotation. Critical hardware includes self-rotating TDM motors, rapid-switching electromagnetic clutches, high-torque worm gear reducers, and dual-encoder sensing systems.

\item \textbf{Robotic arm design.} MuxArm features a disc-spring-backbone configuration enabling compliant multi-segment articulation, achieving 6-DOF control through 9 tendons with only 4 motors while maintaining 2.17 kg self-weight. The disc-spring-bearing assembly (25 bearings providing 44 passive DOFs + 6 active DOFs) offers structural support and passive compliance for high accuracy under actuator failures.

\item \textbf{Trajectory planning.} An actuation space trajectory planning algorithm integrates BeamStep optimization with hardware scheduling, enhancing fault-tolerant control, actuation speed, and reducing tendon loads.

\item \textbf{Experimental validation.}  Experimental validation includes tests of control algorithms and fault tolerance, with successful completion of complex tasks such as obstacle-avoidance grasping and confined-space manipulation, demonstrating its application potential in space exploration and related missions.

\end{itemize}

\section{Related Works}

\begin{table}[b]

\centering
\caption{Summary Comparison of Continuum Robot Driving Solutions}
\label{tab:comparison2}
\scriptsize
\setlength{\tabcolsep}{3pt}
\begin{tabular}{ccccccc}
\hline
\textbf{Work} & 
\textbf{Self-lock} & 
\textbf{Switch Spd.} & 
\textbf{Torque} & 
\textbf{Compact} & 
\textbf{Temp. Sens.} & 
\textbf{Fault Tol.} \\
\hline

Ref.\cite{8746148} & $\times$ & N/A & Medium & Good & Low & $\times$ \\

Ref.\cite{9795861} & \checkmark & 1-3s & Low & Good & High & $\times$ \\

Ref.\cite{7139539} & \checkmark & 1-3s & Low & Good & High & $\times$ \\

Ref.\cite{7487204} & \checkmark & 1-3s & Low & Good & High & $\times$ \\

Ref.\cite{8274862} & \checkmark & 50-80ms & Medium & Poor & Low & $\times$ \\

Ref.\cite{11106755} & \checkmark & 0.1-0.5s & Medium & Good & Low & $\times$ \\

Ref.\cite{11298485} & \checkmark & 0.1-0.5s & High & Poor & Low & $\times$ \\

Ref.\cite{lancaster2024electrostatic} & \checkmark & 50-100ms & Low & Good & Medium & $\times$ \\

\hline

MuxArm & \checkmark & $<$0.1s & High & Good & Low & \checkmark \\

\hline

\end{tabular}
\end{table}

Lightweight and fault-tolerant tendon-driven robots are crucial for practical applications, particularly in aerospace where stringent mass budgets and reliability requirements must be met. Time-division multiplexing (TDM) offers a new paradigm by enabling actuator sharing across DOFs, potentially reducing the actuator-to-DOF ratio without introducing additional mass. Originating from communications engineering, TDM has been applied to tendon-driven systems. Ye et al.~\cite{ye2021time} achieved gait-synchronized path switching in soft exoskeletons, and Zuo et al.~\cite{zuo2020design} used fast-response clutches ($<$100~ms) for sequential joint actuation in 7~mm surgical robots. However, existing TDM implementations have critical limitations (Table~\ref{tab:comparison2}): thermal actuators~\cite{9795861,7139539,7487204} require 1--3~s switching and are temperature-sensitive; electromagnetic~\cite{8274862} and motor-driven~\cite{11298485} systems are bulky; magnetic~\cite{11106755} and electrostatic~\cite{lancaster2024electrostatic} designs lack torque capacity or fault tolerance; base-mounted actuation~\cite{8746148} cannot self-lock. Fundamentally, these employ sequential single-tendon actuation unsuitable for continuum arms requiring coordinated multi-tendon control. Building upon our dexterous hand work~\cite{xu2024muxhand}, this paper extends TDM to continuum arms via a vertically-stacked rotational selection structure and BeamStep algorithm, enabling 1-3 concurrent actuators for coordinated multi-tendon control.

\begin{figure*}[ht]
    \centering
    \includegraphics[width=0.9\linewidth]{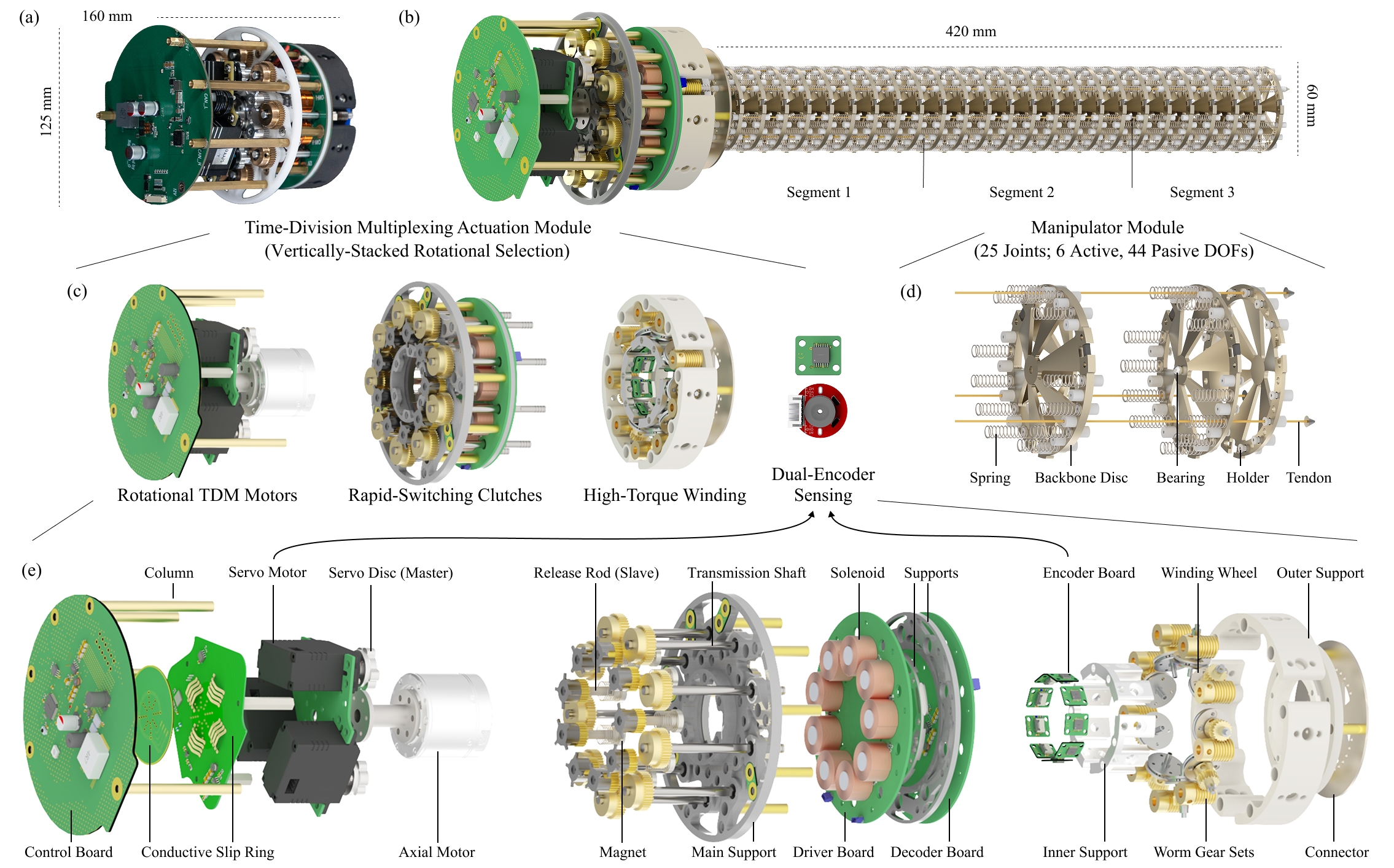}
    \caption{Mechanical design of the MuxArm.
(a) Photograph of TDMA module.
(b) Overall MuxArm assembly, comprising TDMA and manipulator module.
(c) Four critical parts designed for TDMA.
(d) Basic unit of the tendon-driven continuum robotic arm, with multiple such units forming a segment actuated by three tendons for bending.
(e) Exploded view of the driver module: one axial motor repositions three radial servos, sequentially transmitting motion to nine winding wheels in the TDMA scheme.}
    \label{fig:structure}
\end{figure*}

\begin{figure}[h]
    \centering
    \includegraphics[width=\linewidth]{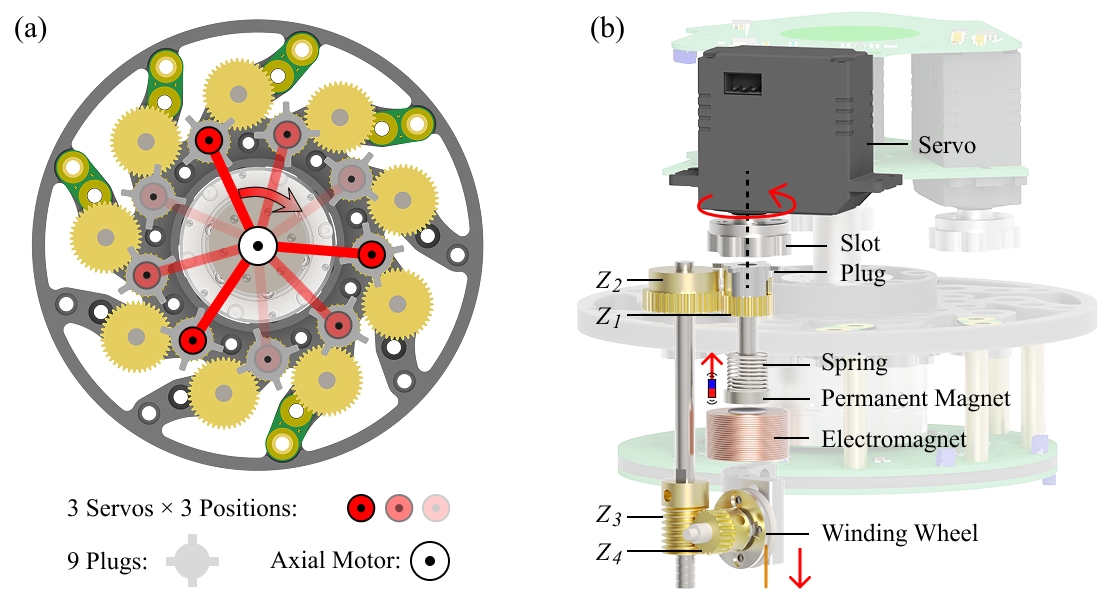}
    \caption{Structural overview of the TDMA scheme.
(a) Cross-sectional view. Axial motor positions three radial servos among nine evenly spaced plugs (shading denotes active positions).
(b) Transmission assembly. The electromagnetic clutch and geartrain transmit servo torque to the winding wheel, with spring-loaded disengagement.}
    \label{fig:tdmm}
\end{figure}

\section{Hardware}

In this section, we present the hardware architecture of the MuxArm, which integrates an innovative driver module and a tendon-driven manipulator to achieve efficient actuation, as shown in Fig.~\ref{fig:structure}(a)-(b). The driver module implements a Time-Division Multiplexing (TDM) scheme with vertically-stacked rotational selection structure as illustrated in Fig.~\ref{fig:structure}(c), including rotational TDM motors, rapid-switching clutch, high-torque winding and dual-encoder sensing.

\subsection{TDMA Operation}

The core concept of TDMA is to manage multiple DOFs by sequentially actuating the radial servos in discrete time slots. The operation includes two main steps: \emph{selecting} and \emph{coupling}. The selecting stage utilizing the axial motor to reposition radial servos and select the tendon, and an electromagnetic clutch couples servo torque to the selected tendon. 

The vertically-stacked rotational selection structure enables compact three-dimensional integration of the actuation system. As is illustrated in Fig.~\ref{fig:tdmm}(a), positioning tasks of three servos are realized through three predefined positions controlled by the TDM motor configuration. The axial motor employs PID loops for both speed and position regulation, thereby accurately positioning the three radial servos. Accordingly, different tendons can be engaged, enabling flexible task execution. In scenarios where servo failure occurs, this mechanism can extend control positions to nine, thus providing redundancy of radial servos.

As in shown in Fig.~\ref{fig:tdmm}(b), torque engagement is facilitated by a electromagnetic clutching mechanism. Specifically, once the servo reaches its designated position, the solenoid will be energized to push the plug on the release rod into the slot on the servo disc. Conversely, when power is disengaged, a spring-loaded mechanism retracts the release rod, disconnecting the torque coupling path. After engagement, the torque is transferred via a gear train, consisting of four gears labeled $Z_1$--$Z_4$, achieving an overall gear reduction ratio of 60. The winding wheel is driven by this amplified torque and its angle is measured by the encoder. Furthermore, unactuated tendon layers are inherently anchored by self-locking worm gears to prevent passive back-driving, while the transient clutch backlash is continuously compensated by the dual-encoder closed-loop system to maintain stable tension.

\subsection{Actuation Module Design}

Leveraging the TDMA scheme, the driver module of MuxArm uses three high-torque radial servos and an axial motor to sequentially actuate nine winding wheels, as shown in Fig.~\ref{fig:structure}(c). A vertically-stacked rotational selection structure is designed to integrate those innovative hardware modules, enabling compact three-dimensional arrangement of the actuation system.

\subsubsection{\textbf{Rotational TDM motors.}} Primarily, the TDM motors with self-rotation capability allows rapid selection of actuator positions, ensuring flexibility and adaptability during operation. One axial motor (HTM4538, Harmonic Drive LLC) is fixed in the center, which drives three radial servos (RX8U50HM, Feetech RC Model Co., Ltd.) placed 120° apart. This vertically-stacked rotational layout is markedly more compact compared to linear design, maximizing space efficiency through three-dimensional integration.
Power and signal communication cables are passed through a conductive slip ring to avoid cable torsion. The CAN protocol is used for signal communication due to its robustness, real-time capability, and noise immunity.

\subsubsection{\textbf{Rapid-Switching Clutches.}} Subsequently, the electromagnetic clutches enable rapid engagement and disengagement within 0.1 seconds, which facilitates swift switching between servos. {The clutch operates via a chamfered cylindrical plug moving into a slot, with pre-engagement motor oscillation, spring-loaded insertion, and dual-encoder verification ensuring reliable engagement. The electromagnetic clutches adopt a normally open design and are energized only during torque transmission, effectively reducing power consumption and temperature rise (approximately 0.45A per coil, temperature around 40℃).
The main support carries the driver board that controls the axial motor and the solenoid. The solenoid is fixed coaxial with the magnets mounted on the release rod. The opposite end of the release rod carries a plug for mating with the servo and a cylindrical gear that couples to one end of the transmission shaft, which pass through linear bearings mounted on the main support.Compared to motors, solenoids exhibit superior durability due to their simple binary operation, which reduces mechanical wear and enhances reliability in aerospace applications.

\subsubsection{\textbf{High-Torque Winding.}} Furthermore, the worm gear reducers increase the winding torque (The force can be transmitted to each tendon by the actuation) and provides self-locking, which maintains stability under power-off conditions and reduces effective reflected inertia at the motor shafts.
The opposite end of each transmission shaft connects to the worm gear sets, which then drive the winding wheel. The nine winding wheels in the actuation module are vertically stacked and used to tension the cables, which in turn bend the MuxArm. The winding wheel is mounted vertically using the inner and outer support. An encoder is installed on the inner side and connects to the decoder board to read the winding wheel angle. The wound tendons exit through the connector and run along the manipulator to their attachment points, where they transmit actuation forces to each joint.

\subsubsection{\textbf{Dual-Encoder Sensing.}} Moreover, the servo encoders and winding wheel encoders are employed to guarantee high positioning accuracy during operation. Highly integrated PCB boards are designed for compact structure. The control board hosts an STM32F446-RET6 microcontroller running at 180 MHz, performing task scheduling, power conversion, and CAN-bus arbitration while tolerating up to 60 V DC. All nodes exchange data over the 1 Mbps CAN backbone. Moreover, it provides a USB serial interface through which high-level kinematic control can be implemented seamlessly. For the winding wheel, each wheel shaft integrates a magnetic encoder (AS5047P) that offers 14-bit resolution (0.02°). Angle data are transferred over SPI at 9 Mbps to the decoder MCU, which forwards the packets to the CAN network.

\subsubsection{\textbf{Lightweight Design}} Finally, several design choices target weight reduction. An apertured plastic–steel main support lowers mass while maintaining strength, and aluminum transmission shafts further reduce inertia. The vertically-stacked rotational structure inherently minimizes the system footprint while enabling efficient three-dimensional space utilization. The copper column between modules are acted simultaneously as structural supports and electrical conductors for passing power and signal. 
This dual-function arrangement reduces cable routing complexity while achieving compact design and ensuring structural stability through lightweight measures, including integrated connectors and hollowed structures.

\subsection{Manipulator Module Design}
As illustrated in Fig.~\ref{fig:structure}(c), the manipulator module of the MuxArm features a lightweight modular design composed of 25 repeating backbone disc units, organized into three segments consisting of 9, 9, and 7 backbone discs, respectively. Each backbone disc consists of two centrally raised Ti–Mg alloy components to achieve high strength while maintaining low weight. A ball bearing accommodates relative angular displacement and interconnects consecutive discs as the central backbone, allowing each segment to achieve multi-directional bending motions. Around each disc pair, twelve sets of symmetrically arranged springs are installed by dedicated holders, providing structural stiffness. Furthermore, three tendon groups, each comprising three evenly spaced tendons, are distributed circumferentially along each backbone disc. These tendons run continuously from winding wheels at the base of the manipulator up to the top of each respective segment, allowing controlled bending along the manipulator length.

\section{Kinematics}
\label{sec:kinematics}

In this section, we derive the forward kinematics (FK) and inverse kinematics (IK) between task, configuration and actuation space~\cite{webster2010design}. MuxArm is modeled as three serially connected continuum segments, each with a constant-curvature~\cite{m1,m2} backbone of fixed length \(L_i\;(i=1,2,3)\), as is depicted in  Fig.~\ref{fig:kinematics}(a) and Table 1. The base frame \(\{T_{0}\}\) is fixed to the mounting flange, whereas \(\{T_{1}\}\), \(\{T_{2}\}\), and \(\{T_{3}\}\) are attached to the distal discs of segments 1, 2, and 3, respectively. For segment \(i\) we adopt the joint coordinates \( \mathbf{q}_i=[\theta_i,\;\phi_i]^\top\), where \(\theta_i\) is the bending angle and \(\phi_i\) is the bending azimuth.  The complete joint vector (configuration space) is \(\mathbf{q}=[\mathbf{q}_1^\top,\mathbf{q}_2^\top,\mathbf{q}_3^\top]^\top\).

\subsection{Task-Configuration Transformation}

Fig.~\ref{fig:kinematics} illustrates the constant curvature modeling for each segment. Eqn.~(1)--(6) follow directly from this model by mapping the segment arc parameters \((\theta,\phi,L)\) to tendon length changes via the routing geometry. For brevity, we omit the intermediate derivation steps.
The yellow centerline is an arc of length \(L\) and radius \(R=L/\theta\), subtending bending angle \(\theta\) in the plane with azimuth \(\phi\). The local frame \(\{T_{n-1}\}\) is fixed to the proximal disc and \(\{T_{n}\}\) to the distal disc. The curvature center \(O_{n}\) lies in the \(x_{n-1}\!-\!y_{n-1}\) plane at distance \(R\) from the backbone. Angle \(\theta\) is measured about \(z_{n-1}\) and \(\phi\) from \(x_{n-1}\) toward \(y_{n-1}\). 

\subsubsection{\textbf{Forward}}
\label{sec:seg_fk}
The homogeneous transformation from \(\{T_{n-1}\}\) to \(\{T_{n}\}\) is:
\begin{equation}
\begin{gathered}
\mathbf{T}(\theta,\phi,L)= \\[-4pt]
\begin{bmatrix}
 c_\phi^{\,2}c_\theta+s_\phi^{\,2} & s_\phi c_\phi(c_\theta-1) & c_\phi s_\theta & \displaystyle \frac{L\,c_\phi(1-c_\theta)}{\theta} \\[4pt]
 s_\phi c_\phi(c_\theta-1)         & c_\phi^{\,2}+s_\phi^{\,2}c_\theta & s_\phi s_\theta & \displaystyle \frac{L\,s_\phi(1-c_\theta)}{\theta} \\[4pt]
 -c_\phi s_\theta                  & -s_\phi s_\theta                 & c_\theta        & \displaystyle \frac{L\,s_\theta}{\theta}           \\[4pt]
 0                                 & 0                                & 0               & 1
\end{bmatrix},
\end{gathered}
\label{eq:seg_fk}
\end{equation}
where \(c_\theta=\cos\theta,\; s_\theta=\sin\theta,\;
      c_\phi=\cos\phi,\;  s_\phi=\sin\phi\). The end effector pose follows from chaining three segments, which forms the forward kinematics:
\begin{equation}
\mathbf{T}_{3}(\mathbf{q})=
\mathbf{T}(\theta_{1},\phi_{1},L_{1})\,
\mathbf{T}(\theta_{2},\phi_{2},L_{2})\,
\mathbf{T}(\theta_{3},\phi_{3},L_{3}),
\label{eq:arm_fk}
\end{equation}
with \(\mathbf{q}\) collecting all joint variables. Eqn.~\eqref{eq:arm_fk} maps the six dimensional configuration space to \(\mathbf{T}_{3}\in\mathrm{SE}(3)\).

\subsubsection{\textbf{Inverse}}
\label{sec:ik}

Given the current and desired pose 
\(\mathbf{T}_\text{u}=
\begin{bmatrix}
\mathbf{R}_\text{u} & \mathbf{p}_\text{u}\\
\mathbf{0}_{1\times3} & 1
\end{bmatrix}\),
\(\mathbf{T}_\text{d}=
\begin{bmatrix}
\mathbf{R}_\text{d} & \mathbf{p}_\text{d}\\
\mathbf{0}_{1\times3} & 1
\end{bmatrix}\),
where \(\mathbf{R}\in\mathrm{SO}(3)\) and \(\mathbf{p}\in\mathbb{R}^3\),
the solution of inverse kinematics uses a damped pseudoinverse iteration.
The \(6\times6\) Jacobian \(\mathbf{J}(\mathbf{q})\) is assembled recursively from segment level angular and linear blocks. At each iteration, the pose error \(\boldsymbol{\varepsilon} = \begin{bmatrix} \mathbf{p}_{\text{d}} - \mathbf{p}_{\text{u}} \\ \frac{1}{2}(\mathbf{R}_{\text{u}}^{\top}\mathbf{R}_{\text{d}} - \mathbf{R}_{\text{d}}^{\top}\mathbf{R}_{\text{u}})^{\vee} \end{bmatrix}\)~\cite{blanco2021tutorial} is projected through \(\mathbf{J}^{\dagger}=\mathbf{J}^{\!\top}(\mathbf{J}\mathbf{J}^{\!\top})^{-1}\) to obtain:
\begin{equation}
\Delta\mathbf{q}= \lambda\,\mathbf{J}^{\dagger}\boldsymbol{\varepsilon},
\label{eq:ik_update}
\end{equation}
with gain \(\lambda\) initialized at 1 and halved on non-descent steps.
Joint limits on \(\theta_i\) are enforced after each update, and convergence is declared when position and orientation errors fall below \(10^{-10}\).

To enhance robustness and avoid local minima, a null space term \(\Delta\mathbf{q}_{\mathrm{null}}=(\mathbf{I}-\mathbf{J}^{\dagger}\mathbf{J})\boldsymbol{\mu}\) resolves redundancy, where $\boldsymbol{\mu}$ denotes a null-space bias. Moreover, a parallel multi start scheme launches many local solves randomly, retaining the best collision free configuration for full pose only objectives.

\begin{figure}[t]
    \centering
    \includegraphics[width=0.9\linewidth]{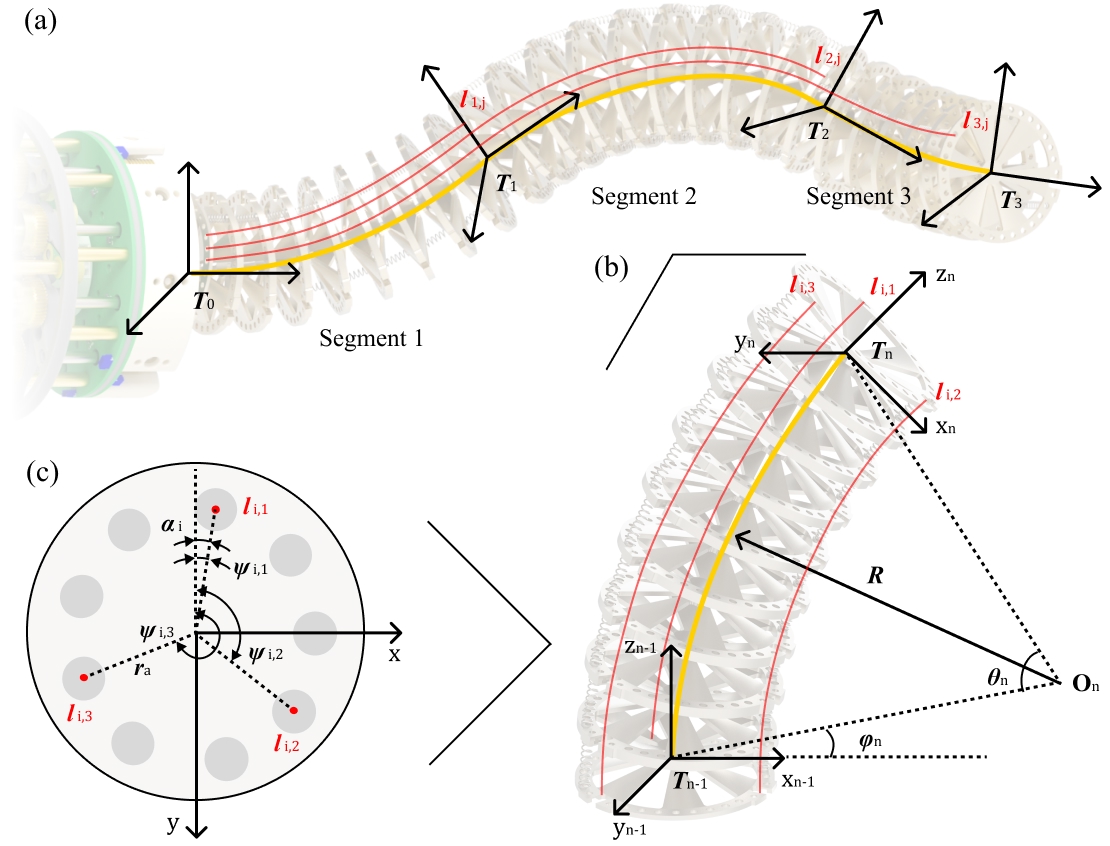}
    \caption{Hierarchical kinematic representation of the MuxArm.  
(a) Complete backbone assembly of three serial segments.  
(b) Individual segment model with constant curvature.  
(c) Cross-sectional view of the backbone unit.}

    \label{fig:kinematics}
\end{figure}

\subsection{Configuration--Actuation Transformation}
\label{sec:jt_map}

Fig.~\ref{fig:kinematics}(a) highlights each segment \(i\in\{1,2,3\}\) driven by three tendons sets in red. These tendons follow the yellow backbone arc and attach to every backbone disc, as shown in Fig.~\ref{fig:kinematics}(b). A cross section in Fig.~\ref{fig:kinematics}(c) shows three anchor points at a common radius \(r_\text{a}\), each separated by 120\(^\circ\). The tendons are indexed by \(j\in\{1,2,3\}\). In total, nine tendons are evenly distributed to match the nine winding wheels, although only three are marked in the cross-section. The tendon azimuth is defined as \(\psi_{i,j}= \alpha_i + \frac{2\pi}{3}(j-1)\), where \(\alpha_i\) is the base angular offset (segment \(i\)'s tendon-triad orientation) and \(\psi_{i,j}\) is the azimuthal angle of tendon \(j\) within segment \(i\).
\subsubsection{\textbf{Forward}} Let \(\Delta\mathbf{l}_i\) be measured tendon displacements. First compute an intermediate azimuth \(\tilde{\phi}_i\) in the local disc frame:
Based on the discrete constant-curvature framework, the bending parameters can be geometrically decoupled. We first compute an intermediate azimuth \(\tilde{\phi}_i\) in the local disc frame:
\begin{equation}
\tilde{\phi}_i=
\tan^{-1}\!
\frac{\sqrt{3}\,(\Delta l_{i,2}-\Delta l_{i,3})}
     {2\Delta l_{i,1}-\Delta l_{i,2}-\Delta l_{i,3}},
     \label{eq:tendon_ik}
\end{equation}
and the global bending azimuth is \(\phi_i=\tilde{\phi}_i+\alpha_i\; (\text{mod }2\pi)\).  Subsequently, by substituting \(\tilde{\phi}_i\) into the geometric projection and utilizing the sum of the tendon displacements to isolate the bending magnitude, it follows as:
\begin{equation}
\theta_i =
\frac{\sqrt{3(\Delta l_{i,2}{-}\Delta l_{i,3})^{2}
+3(\Delta l_{i,2}{+}\Delta l_{i,3}{-}2\Delta l_{i,1})^{2}}\,L_i}
     {r_\text{a}\sqrt{1+2\cos^{2}\tilde{\phi}_i}\,
      \bigl[3L_i{+}(\Delta l_{i,1}{+}\Delta l_{i,2}{+}\Delta l_{i,3})\bigr]}.
\end{equation}

\subsubsection{\textbf{Inverse}}
The length change of tendon \(j\) in segment \(i\) is:
\begin{equation}
\!\Delta l_{i,j}\!=\!
\sum_{n=1}^{i}\!
\left[
2\xi\sin\!\left(\frac{\theta_n}{2\xi}\right)\!\!
\left(\!\frac{L_n}{\theta_n}\!-\!r_\text{a}\!\cos(\psi_{n,j}\!-\!\phi_n)\!\!\right)\!
\!-\!L_n
\right],
\label{eq:tendon_fk}
\end{equation}
where \(\xi\) is the number of backbone discs per segment. The tendons pass through discrete guides on these discs, so the tendon path is modeled as \(\xi\) straight segments, while the backbone still follows a constant-curvature arc. The tendons in the same section are denoted as \(\Delta\mathbf{l}_i= [\Delta l_{i,1},\Delta l_{i,2},\Delta l_{i,3}]^\top\).

Eqn.~\eqref{eq:tendon_ik}-\eqref{eq:tendon_fk} provide a bijective mapping between configuration space \(\mathbf{q}=[\theta_1,\phi_1,\theta_2,\phi_2,\theta_3,\phi_3]^\top\) and actuation space \(\Delta\mathbf{l}=[\Delta\mathbf{l}_1^\top,\Delta\mathbf{l}_2^\top, \Delta\mathbf{l}_3^\top]^\top\), forming the analytical bridge between the actuation space planner and the hardware actuation strategy.

\section{TDMA Algorithm}
\label{sec:algorithm}

This section outlines the algorithmic framework of TDMA for MuxArm. As illustrated in Fig.~\ref{fig:frameworks}, environmental sensing provides the obstacle map and desired end-effector pose in task space. The IK solver converts this into an optimal joint configuration, which is then mapped into actuation space. We formulate tendon trajectory planning as a mixed-integer optimization problem, solved incrementally using the BeamStep method to produce discrete tendon length update sequences. A time-division multiplexing scheduler converts these updates into executable motor commands respecting hardware transmission constraints. The command sequence is delivered to the onboard STM32 controller, driving the axial motor, radial servos, and electromagnets to realize planned motion. Encoders on the winding wheels provide feedback for closed-loop control.

\begin{figure}[t]
    \centering
    \includegraphics[width=0.9\linewidth]{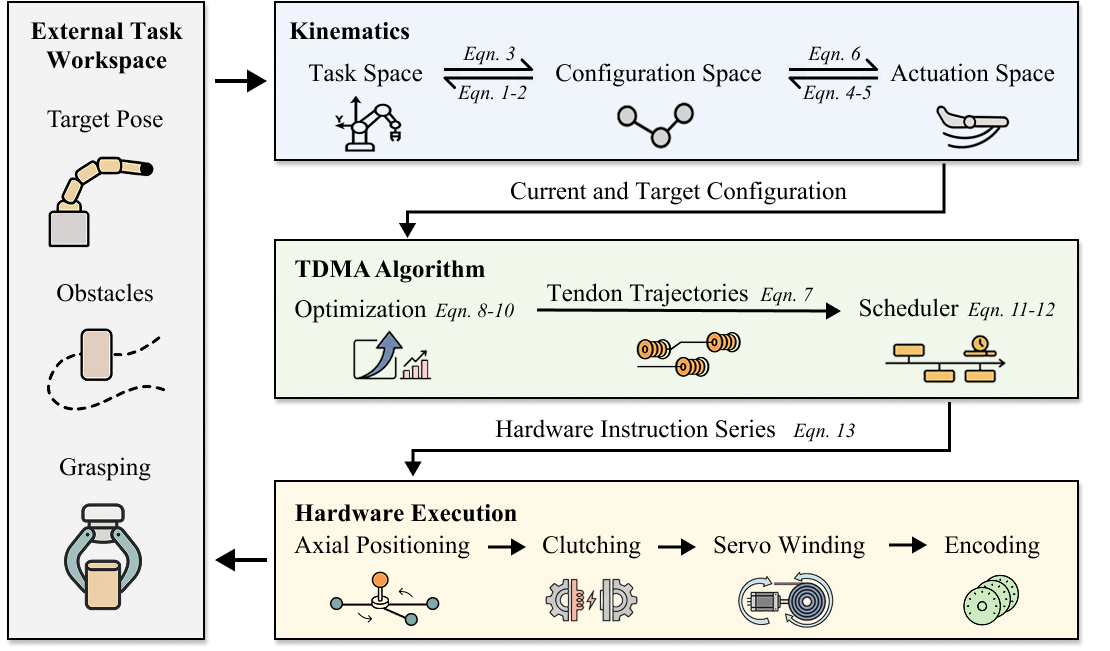}
    \caption{Control framework of the MuxArm. Environmental inputs are mapped to actuation space trajectories, then optimized by the TDMA algorithm into actuator commands, and finally executed by an onboard STM32 controller.}
    \label{fig:frameworks}
\end{figure}

\subsection{Optimization}
The trajectory planning problem aims to determine a sequence of incremental adjustments of the tendons subject to movement constraints. The manipulator is actuated by nine tendons with lengths $\mathbf{l}=[l_1,\dots,l_9]^\top$.\footnote{To simplify notation in this section, we use flattened single subscripts \(l_c\) instead of the double-index form \(l_{i,j}\) for the the nine individual tendons.} Given an initial state $\mathbf{l}^0$ and a target state $\mathbf{l}^\star$ in the actuation space, the trajectory is modeled as a piecewise‐linear path $\{\mathbf{l}^k\}_{k=0}^K$ with $K$ steps. Each increment lies in the span of the standard basis vectors indexed by hardware availability $\mathcal{I}^k$:
\begin{equation}
\Delta\mathbf{l}^k \; = \sum_{c\in\mathcal{I}^k} \Delta l^k_c\,\mathbf{e}_c \; \in\; \operatorname{span}\{\mathbf{e}_c : c \in \mathcal{I}^k\},
\end{equation}
where $\mathbf{e}_c$ denotes standard basis vector in $\mathbb{R}^{9}$, and $\mathcal{I}^k$ can be stated as the layer constraint. Let the tendon indices be grouped as $P_1=\{1,2,3\}$, $P_2=\{4,5,6\}$, $P_3=\{7,8,9\}$ according to the TDMA design. At step $k$, motions are restricted to a single active layer $P_{i^k}$ determined by the available servos $s \in \Sigma \subseteq \{1,2,3\}$. And only the one tendon group associated with a single segment is actuated, while the other two groups are held constant in the actuation space. The available tendon set $\mathcal{I}^k$ must lie within that layer:
\begin{itemize}
    \item $|s|=1$: $|\mathcal{I}^k|=1$, with $\mathcal{I}^k\subseteq P_{i^k}$;
    \item $|s|=2$: $|\mathcal{I}^k|=2$, both in the same layer, $\mathcal{I}^k\subseteq P_{i^k}$;
    \item $|s|=3$: $|\mathcal{I}^k|=3$, the whole layer, $\mathcal{I}^k=P_{i^k}$.
\end{itemize}

Additionally, we add a step size constrain. Per–step scalar increments $\Delta l^k_c$ are clipped with step size $\Delta_{\max}$. For the cost function, we formulates travel distance, bending energy, and motor switches of the manipulator as: 
\begin{equation}\label{equ:cost}
\mathcal{J}
=\sum_{k=1}^K\sum_{c\in\mathcal{I}^k}\bigl|\Delta l^k_{c}\bigr|
\;+\;\lambda_\text{b}\sum_{k=1}^K\|\boldsymbol{\theta}(\mathbf{l}^k)\|^2
\;+\;\lambda_\text{sw}\,W_\text{sw},
\end{equation}
where $\mathbf{l}^k=\mathbf{l}^0+\sum_{n=1}^k\Delta\mathbf{l}^n$ is the intermediate state, $\boldsymbol{\theta}(\mathbf{l}^k)$ is the forward kinematics of the pose angles, and $\lambda_\text{b},\lambda_\text{sw}$ are tunable weights, $W_\text{sw}$ is the switches count of axial motor. In this way, trajectory planning is posed as the following optimization:
\begin{subequations}
\begin{align}
&\min_{\{\mathcal{I}^k,\Delta\mathbf{l}^k\}_{k=1}^K}\;\mathcal{J},\\
\text{s.t.}\;&|\Delta l^k_{c}|\le\Delta_{\max},\ \forall\,c,k,\\
&\mathcal{I}^k\subseteq P_{i^k},\ \forall\,k,
\end{align}
\end{subequations}
where $K$ is a also decision variable and constraints encode the layer restriction and the per step size bound stated above. 

\begin{algorithm}[b]
\caption{BeamStep.}\label{alg:BeamStep}
\begin{algorithmic}
\STATE {\textsc{BeamStep}}$(\mathbf{l}^0, \mathbf{l}^\star, \Delta_{\max}, \lambda_\text{b}, \lambda_\text{w}, B, \mathrm{tol})$
\STATE $H \gets$ min-heap with $(\mathbf{l}^0,\ \mathcal{J}=0)$
\WHILE{$H$ not empty}
  \STATE Keep top $B$ in $H$
  \STATE $(\mathbf{l},\mathcal{J}) \gets$ pop lowest-cost from $H$
  \IF{$\|\mathbf{l}-\mathbf{l}^\star\|_\infty \le \mathrm{tol}$}
    \STATE \textbf{return} reconstructed trajectory
  \ENDIF
  \FOR{each valid tendon set $\mathcal{I}^k$}
    \STATE Compute $\Delta\mathbf{l}^k$ 
    \STATE Evaluate cost contribution using Eqn.~\eqref{equ:cost}
    \STATE Insert $(\mathbf{l}+\Delta\mathbf{l}^k,\ \text{updated }\mathcal{J})$ into $H$
  \ENDFOR
\ENDWHILE
\end{algorithmic}
\end{algorithm}

We adopt the BeamStep procedure to solve this mixed‐integer optimization problem, due to the discrete sets $\mathcal{I}^k$ and continuous variables $\Delta l^k_c$. It utilises the Beam search~\cite{beam} that keeps the $B$ lowest-cost partial trajectories at each expansion with step constraint $\Delta_{\max}$. Starting from $\mathbf{l}^0$, the algorithm repeatedly enumerates feasible $\mathcal{I}^k\subseteq P_{i^k}$ to obtain $\Delta\mathbf{l}^k$, (iii) evaluates the cost in Eqn.~\eqref{equ:cost} and keeps the best $B$ candidates in a min-heap, and (iv) advances until $\|\mathbf{l}^k-\mathbf{l}^\star\|_\infty\le\mathrm{tol}$. The retained increments are then chained to produce the executable sequence $\mathbf{l}^0,\dots,\mathbf{l}^K$. Algorithm~\ref{alg:BeamStep} summarizes the procedure.

\subsection{Scheduler}
\label{sec:scheduler}

To execute tendon trajectories on hardware, incremental tendon motions in actuation space must be translated into servo commands that respect the TDMA scheme. We describe the scheduler in three parts: (1) index mapping among tendons, axial positions and radial servos, (2) fusion of consecutive motions, and (3) the final plan.

\subsubsection{\textbf{Mapping}}
Let $a \in \mathcal{A} = \{1,\dots,9\}$ be the axial positions. With the tendons as $l_c$, the TDMA scheme can be modeled as a surjective deterministic map \(c = F(a,s)\). At step $k$, each required tendon $l_c^k$ is realized by choosing:
\begin{equation}
(a^k,s^k)\in F^{-1}(c^k),
\label{eq:mapping}
\end{equation}
where $F^{-1}(l)=\{(a,s)\mid F(a,s)=l\}$. Among all feasible selections, the scheduler chooses the one that minimizes expected axial motor switches (e.g., by reusing $a^{k-1}$) and balances future TDMA constraints. In this way, each commanded tendon $l_c$ is converted into a concrete pair $(a,s)$.

\subsubsection{\textbf{Fusion}}
Consecutive slices that use the same axial position are accumulated and merged. After this fusion, the original $K$ actuation steps are reduced to $K_\text{h}$ hardware stages ($K_\text{h}\le K$). For axial position $a$ and servo $s$, the fused tendon displacement is:
\begin{equation}
\Delta l^{*}(a,s)=\sum_{k:\,a^k=a^{k-1}}\Delta l^{k}_{F({a,s})},
\label{eq:fused}
\end{equation}
where commands with $|\Delta l^{*}(a,s)|<\text{tol}$ are discarded.

\subsubsection{\textbf{Final plan}}
After fusion, the net tendon displacements $\Delta l^*(a,s)$ could be converted into servo rotation angles $\Delta\beta_{s}$ by a scale factor $N$ determined by the transmission ratio, i.e., $\Delta\beta_{s} = N\Delta l^{*}(a,s)$. Moreover, each clutch $g$ corresponds one-to-one with its actuated tendon $l_c$. The scheduler then emits an ordered list of $K_\text{h}$ hardware instructions:
\begin{equation}
\bigl\{(a^k,\;g^k,\;[\,(s^k,\Delta\beta^k_{s})\,]_{s\in\Sigma})\bigr\}_{k=1}^{K_\text{h}},
\label{eq:plan}
\end{equation}
where each instruction includes the axial position $a^k$, the clutch $g^k$ to engage, and the commanded angle $\Delta\beta^k_{s}$ for servo $s$. This plan is compact because axial motor switches are minimized and all motions at the same axial position are fused before streaming to the STM32 controller.

\section{Experiments}

In this section, first, we benchmark TDMA through trajectory analysis and statistical study. Second, we assess fault tolerance by emulating servo failures with optical motion tracking. MuxArm's performance is quantified via load tests and workspace validation. Finally, we demonstrate manipulation in cluttered and confined settings reflecting on-orbit servicing scenarios like debris capture and equipment access.

\begin{figure*}[t]
    \centering
    \includegraphics[width=0.9\linewidth]{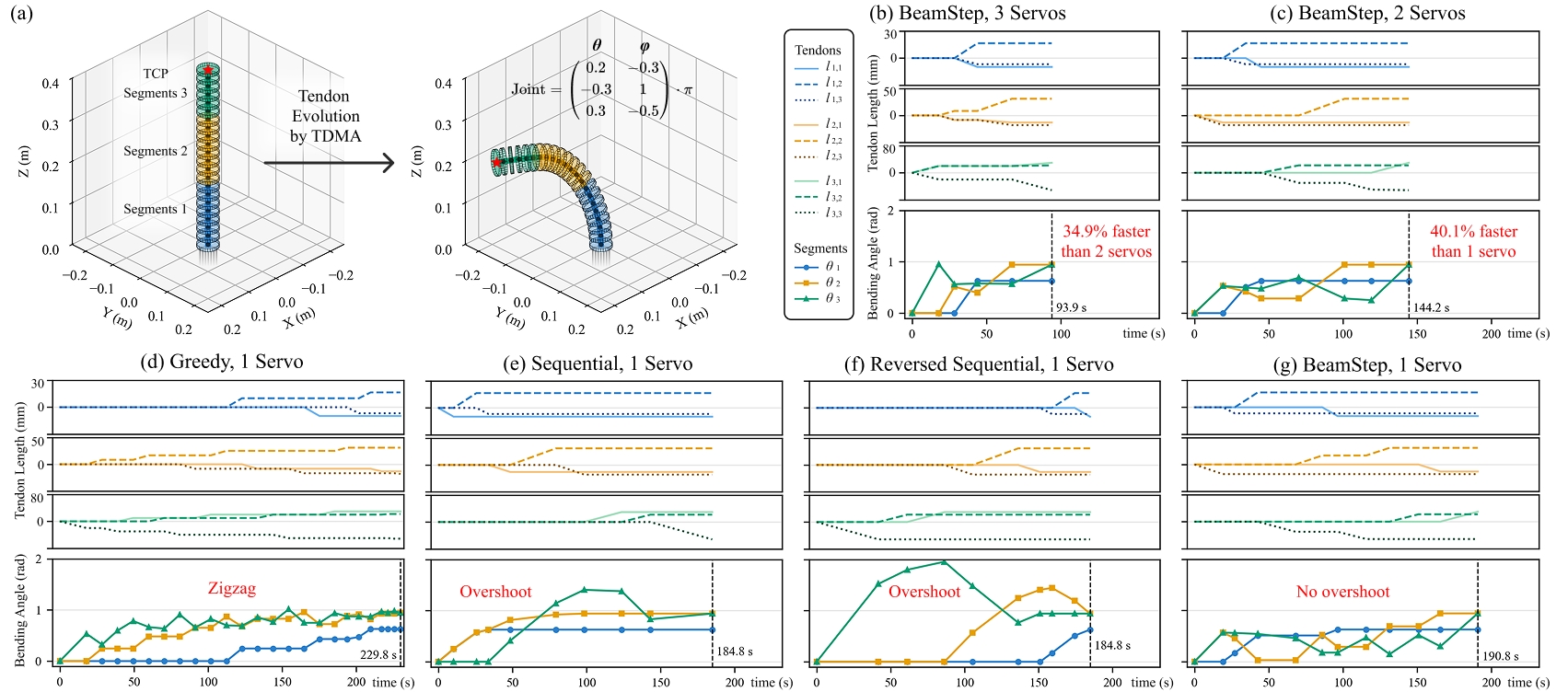}
    \caption{Experimental validation of TDMA-based trajectory planning for MuxArm, evaluated by tendon length and bending angle variation during the evolution process.
(a) Illustration of the initial and target configuration of MuxArm.
(b), (c), and (g) Results of the proposed BeamStep method under different numbers of available servos (1–3).
(d), (e), and (f) Results of three baseline methods with one servo available.}
    \label{fig:procedure}
\end{figure*}

\subsection{Algorithm Validation}
We benchmark the TDMA actuation space planner on MuxArm against three baseline policies. For segment $i\in\{1,2,3\}$ with joint coordinates $\mathbf{q}_i=[\theta_i,\phi_i]^\top$, the tendon length changes are $\Delta\mathbf{l}_i=[\Delta l_{i,1},\Delta l_{i,2},\Delta l_{i,3}]^\top$.

\subsubsection{\textbf{Baselines}}
Let $\pi(k)$ be the policy to select the tendon actuated at step $k$. The update applies only to the selected tendon; others are zero. We consider three policies: a sequential scan cycling 1$\to$9 (Eqn.~\eqref{eq:baseline:seq}), a reverse scan cycling 9$\to$1 (Eqn.~\eqref{eq:baseline:rev}), and a greedy rule that selects the tendon with the largest absolute residual to its target (Eqn.~\eqref{eq:baseline:greedy}). The policies are as follows:\begin{subequations}
\label{eq:baseline}
\begin{align}
\pi_{\text{seq}}(k) &= 1 + \bigl((k-1)\bmod 9\bigr), \label{eq:baseline:seq}\\
\pi_{\text{rev}}(k) &= 9 - \bigl((k-1)\bmod 9\bigr), \label{eq:baseline:rev}\\
\pi_{\text{greedy}}(k,\Delta_{\max}) &= \arg\max_{(i,j)} \bigl|\,l_{i,j}^* - l_{i,j}^{k-1}\,\bigr|, \label{eq:baseline:greedy}
\end{align}
\end{subequations}
after which, the scheduler converts the net tendon displacements into servo commands according to Eqn.~\eqref{eq:plan}.

\subsubsection{\textbf{Metrics}}
We evaluate each plan by execution time $T$ and transient quality $\rho^{\text{duration}}_i$, and $\rho^{\text{peak}}_i$. During hardware instruction $k$, engaged servos are synchronized with proportionally scaled speed under a global speed limit $\omega$. The motion time and total time are:
\begin{equation}
t_{\text{move}}^{k}=\max_{s\in\Sigma^k}\frac{|\Delta\beta^{k}_{s}|}{\omega},\qquad
T=\sum_{k=1}^{K_\text{h}}\bigl(t_{\text{sw}}+t_{\text{move}}^{k}\bigr),
\label{eq:time}
\end{equation}
where $t_{\text{sw}}$ denotes the time of repositioning and engagement. 

For each segment $i\in\{1,2,3\}$ with bending angle $\theta_i(t)$ and steady value $\theta_i^{\infty}$, we report two overshoot ratios:
\begin{subequations}
\begin{align}
&\rho^{\text{duration}}_i = \frac{1}{T}\int_{0}^{T}\mathbf{1}\!\left\{\,|\theta_i(t)|>|\theta_i^{\infty}|\,\right\}\,dt, \label{eq:time_ratio}\\
&\rho^{\text{peak}}_i= \frac{\displaystyle\max_{t\in[0,T]}\bigl(|\theta_i(t)|-|\theta_i^{\infty}|\bigr)_{+}}{|\theta_i^{\infty}|},(x)_{+}=\max(x,0).
\label{eq:mag_ratio}
\end{align}
\end{subequations}

\begin{table*}[t]
  \centering
  \caption{Comparison of different planning methods across various servo count. Values are reported as Median (IQR).}
  \label{tab:servo-comparison}
    \renewcommand{\arraystretch}{1} 
    \resizebox{1.9\columnwidth}{!}{
  \begin{tabular}{cc*{4}{cc}}
    \toprule
    & & \multicolumn{2}{c}{\textbf{BeamStep~(Ours)}}
        & \multicolumn{2}{c}{\textbf{Reversed Sequential}}
        & \multicolumn{2}{c}{\textbf{Sequential}}
        & \multicolumn{2}{c}{\textbf{Greedy}} \\
    \midrule
    \multirow{6}{*}{\makecell{One\\Servo}}
      & Total Time [s]   & \multicolumn{2}{c}{\textbf{114.6~(41.8)}}
                   & \multicolumn{2}{c}{114.6~(35.8)}
                   & \multicolumn{2}{c}{114.6~(35.8)}
                   & \multicolumn{2}{c}{141.6~(48.0)} \\
      & Total Stage  & \multicolumn{2}{c}{\textbf{10.0~(2.5)}}
                   & \multicolumn{2}{c}{10.0~(0.0)}
                   & \multicolumn{2}{c}{10.0~(0.0)}
                   & \multicolumn{2}{c}{17.0~(4.5)} \\
      \cmidrule(lr){3-4} \cmidrule(lr){5-6}
      \cmidrule(lr){7-8} \cmidrule(lr){9-10}
      & Overshoot  & Duration [\%] & Peak [\%]
                   & Duration [\%] & Peak [\%]
                   & Duration [\%] & Peak [\%]
                   & Duration [\%] & Peak [\%]\\
      \cmidrule(lr){3-4} \cmidrule(lr){5-6}
      \cmidrule(lr){7-8} \cmidrule(lr){9-10}
      & Segment 1        & \textbf{0.0~(0.0)}   & \textbf{0.0~(0.0)}
                   & 0.0~(0.0)   & 0.0~(0.0)
                   & 0.0~(0.0)   & 0.0~(0.0)
                   & 0.0~(0.0)   & 0.0~(0.0) \\
      & Segment 2        & \textbf{0.0~(7.3)}   & \textbf{0.0~(24.3)}
                   & 20.5~(33.4) & 39.2~(72.1)
                   & 23.2~(41.3) & 21.3~(117.0)
                   & 19.3~(13.3) & 14.1~(24.7) \\
      & Segment 3        & \textbf{9.5~(25.1)}  & \textbf{0.0~(19.0)}
                   & 13.0~(49.6) & 14.7~(91.6)
                   & 70.2~(28.0) & 49.9~(55.6)
                   & 25.2~(26.6) & 14.7~(21.0) \\
    \midrule
    \multirow{6}{*}{\makecell{Two\\Servos}}
      &Total Time [s]   & \multicolumn{2}{c}{96.2~(23.1)}
                   & \multicolumn{2}{c}{81.3~(20.8)}
                   & \multicolumn{2}{c}{79.8~(20.3)}
                   & \multicolumn{2}{c}{74.2~(26.0)} \\
      & Total Stage  & \multicolumn{2}{c}{\textbf{7.0~(1.0)}}
                   & \multicolumn{2}{c}{8.0~(1.0)}
                   & \multicolumn{2}{c}{7.0~(0.0)}
                   & \multicolumn{2}{c}{10.0~(2.0)} \\
      \cmidrule(lr){3-4} \cmidrule(lr){5-6}
      \cmidrule(lr){7-8} \cmidrule(lr){9-10}
      & Overshoot    & Duration [\%] & Peak [\%]
                   & Duration [\%] & Peak [\%]
                   & Duration [\%] & Peak [\%]
                   & Duration [\%] & Peak [\%]\\
      \cmidrule(lr){3-4} \cmidrule(lr){5-6}
      \cmidrule(lr){7-8} \cmidrule(lr){9-10}
      & Segment 1        & \textbf{0.0~(0.0)}   & \textbf{0.0~(0.0)}
                   & 0.0~(0.0)   & 0.0~(0.0)
                   & 0.0~(0.0)   & 0.0~(0.0)
                   & 0.0~(0.0)   & 0.0~(0.0) \\
      & Segment 2        & \textbf{22.2~(46.0)} & \textbf{15.8~(76.2)}
                   & 23.5~(36.3) & 28.7~(66.9)
                   & 28.6~(35.8) & 21.3~(118.8)
                   & 24.4~(25.0) & 18.4~(19.5) \\
      & Segment 3        & 28.0~(49.0) & \textbf{0.0~(31.2)}
                   & 19.3~(52.2) & 11.1~(91.6)
                   & 67.4~(20.1) & 49.9~(59.4)
                   & 23.7~(30.9) & 18.5~(20.1) \\
    \midrule
    \multirow{6}{*}{\makecell{Three\\Servos}}
      & Total Time [s]  & \multicolumn{2}{c}{55.8~(18.5)}
                   & \multicolumn{2}{c}{52.8~(17.9)}
                   & \multicolumn{2}{c}{52.8~(17.9)}
                   & \multicolumn{2}{c}{64.8~(23.9)} \\
      & Total Stage  & \multicolumn{2}{c}{5.0~(1.0)}
                   & \multicolumn{2}{c}{4.0~(0.0)}
                   & \multicolumn{2}{c}{4.0~(0.0)}
                   & \multicolumn{2}{c}{8.0~(2.0)} \\
      \cmidrule(lr){3-4} \cmidrule(lr){5-6}
      \cmidrule(lr){7-8} \cmidrule(lr){9-10}
      & Overshoot    & Duration [\%] & Peak [\%]
                   & Duration [\%] & Peak [\%]
                   & Duration [\%] & Peak [\%]
                   & Duration [\%] & Peak [\%]\\
      \cmidrule(lr){3-4} \cmidrule(lr){5-6}
      \cmidrule(lr){7-8} \cmidrule(lr){9-10}
      & Segment 1        & \textbf{0.0~(0.0)}   & \textbf{0.0~(0.0)}
                   & 0.0~(0.0)   & 0.0~(0.0)
                   & 0.0~(0.0)   & 0.0~(0.0)
                   & 0.0~(0.0)   & 0.0~(0.0) \\
      & Segment 2        & \textbf{7.4~(28.6)}  & \textbf{1.7~(18.5)}
                   & 26.3~(46.5) & 28.7~(67.3)
                   & 29.8~(37.4) & 15.8~(118.8)
                   & 24.2~(19.8) & 16.1~(38.2) \\
      & Segment 3        & 28.9~(40.3) & 14.1~(23.9)
                   & 23.9~(50.8) & 11.1~(91.6)
                   & 55.6~(16.1) & 49.9~(59.4)
                   & 38.8~(32.7) & 21.8~(44.4) \\
    \bottomrule
  \end{tabular}
  }
\end{table*}

\subsubsection{\textbf{Single-trajectory case study}}
Fig.~\ref{fig:procedure}(a) illustrates the motion from the straight state to the target $\mathbf{q}^{*}$, with the three segments colour-coded and the TCP marked. BeamStep combines high performance with efficiency. Panels (b), (c), and (g) show BeamStep with the scheduler for 3, 2, and 1 available servos. Between three and two servos, the time shortens by 34.9\%, which is smaller than the 50\% difference in actuator count. The same holds between one and two servos with a 40.1\% speed gain. This will be further supported by the statistical study and the fault tolerance evaluation.

BeamStep suppresses overshoot and reduces execution time. Panels (d)–(f) present the baselines with one servo. Greedy requires the longest time and induces zigzag tendon updates. Sequential and reversed sequential complete the fastest among the baselines; however, the former generates large intermediate angles on the distal segment, while the latter shows even more pronounced overshoot. These patterns indicate higher transient tendon load. In contrast, BeamStep exhibits a smooth and shortened tendon evolution, reducing tension by up to 50\% with suppressed overshoot.

\subsubsection{\textbf{Statistical analysis across tasks}}
We evaluate 15 randomly sampled feasible targets. For each method and for one to three available servos, we report the median and interquartile range of total hardware stages $K_\text{h}$, total time $T$ and the per–segment overshoot ratios $\rho_i$. Table~\ref{tab:servo-comparison} summarises the results (best in bold). Qualitatively, the control strategy tends to use the proximal layer for coarse positioning to reduce switching, and reserves distal-layer updates for late-stage orientation refinement. Moreover, including the switching penalty discourages frequent layer changes; without this term, the behavior approaches the Greedy baseline with noticeably more axial toggling and zigzag tendon updates.

BeamStep also produces smaller transients on all segments. With one servo, overshoot on segment~1 is consistently negligible because it is closest to the base and most constrained by the structure, while segments~2--3 show larger transients that decay as available servos increase; In contrast, the baselines show longer durations and larger peaks with medians often above 50\% and peaks that can exceed the steady value, especially on the distal segment. This follows from respecting tendon coupling and ordering updates to avoid antagonistic interactions, which lowers transient tendon load. 

\begin{figure*}[t]
    \centering
    \includegraphics[width=0.95\linewidth]{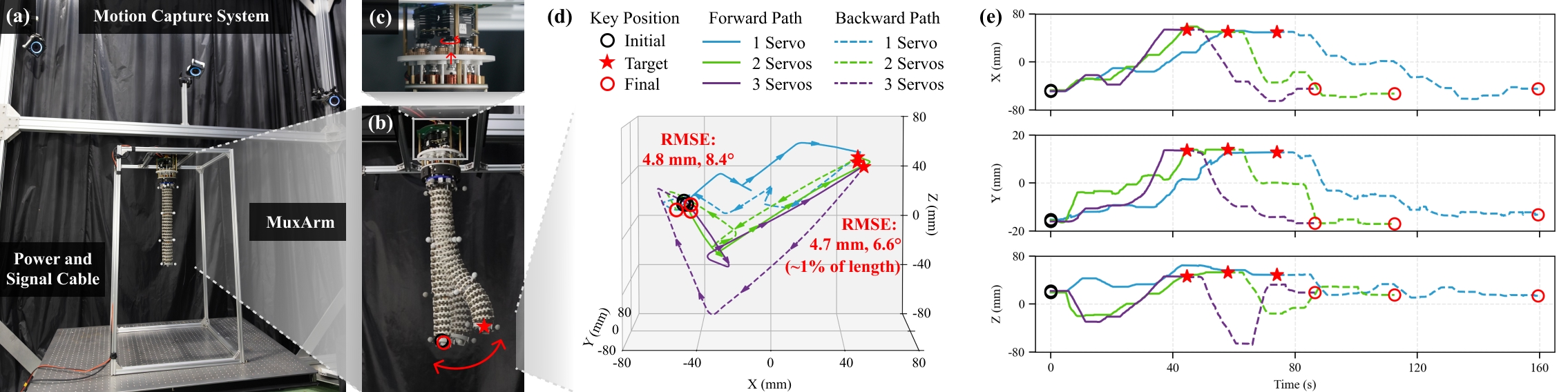}
    \caption{Validating fault tolerance control of MuxArm. (a) Experimental setup with motion capture system. (b) Snapshots of the MuxArm’s forward and backward movement between the relaxed state and the target.
(c) Clutching of the actuation module.
(d) End-effector trajectories in task space under different numbers of active servos.
(e) Decomposed trajectories over time, demonstrating that all configurations achieve the same target pose with high accuracy.}
    \label{fig:mocap}
\end{figure*}

The statistics also confirm the performance and efficiency. As available servos decrease, BeamStep’s median operational time grows less than proportionally, with one–servo times well below three times of the three–servo reference. In contrast, baselines require more hardware stages or suffer longer transients. This shows that under time-division scheduling, our hardware and algorithm deliver high performance with multiple motors and maintain efficiency with few motors. The minimized total execution time effectively demonstrates the prevention of geometrically inefficient paths. Given the time-multiplexed nature of TDMA, it concurrently serves as an effective indicator to validate low task-space deviations.

\subsection{Fault Tolerance}

Our fault-tolerance approach primarily addresses winding servo failures. While the axial motor represents a single point of failure for layer switching, its significantly lower load, duty cycle, and use of brushless motor results in higher reliability.
To evaluate the fault tolerance by TDMA, we emulate partial servo failures on MuxArm and track closed-loop motion. As shown in Fig.~\ref{fig:mocap}(a), experiments run inside a motion capture system (OptiTrack V120 Trio) at $120\,\mathrm{Hz}$ with rigidly mounted MuxArm. Four marker triads are attached to the base and to the distal end of each section to estimate key-point poses (Fig.~\ref{fig:mocap}(b)). Each trial starts from the straight configuration, reaches a target, and returns. We repeat the protocol with one, two, and three active servos.

Across five independent trials and all failure conditions (3/2/1 active servos), the mean target RMSE remained low at \(4.6\,\mathrm{mm}\) in position and \(7.2^\circ\) in orientation, indicating only minor accuracy degradation under partial actuation. We report one representative trial. Fig.~\ref{fig:mocap}(c) highlights the selecting and clutching hardware. In Fig.~\ref{fig:mocap}(d)–(e), solid curves denote forward motion, dashed curves denote the return, and colour encodes the number of available servos. From the capture data, the position RMSE at the target is 4.7 mm, about 1\% of the arm length, and the orientation is 6.6$^\circ$, while it's slightly higher over the full motion. The key positions marked in Fig.~\ref{fig:mocap}(d) show nearly coincident targets, indicating that TDMA maintains high accuracy under faults.

For the result, the accuracy remains high because MuxArm retains full-DOF actuation. The source of error is the clutch slip and coupling mismatch during axial switches, which are amplified by the gear train and result in approximately $0.5\,\mathrm{mm}$ of tendon length change per switch. Fewer servos imply more switches, yet the cumulative error remains modest, which explains the tight clustering in Fig.~\ref{fig:mocap}(d)–(e). These observations verify intrinsic hardware robustness and effective redundant control under partial servo failures.

\subsection{Load Test \& Workspace Validation}

To substantiate the enhanced performance enabled by TDMA, we conduct payload tests and workspace verification. Fig.~\ref{fig:load}(a) shows the self-weight of MuxArm as 2.17 kg. In Fig.~\ref{fig:load}(b), the driver module sustains a 10 kg payload, achieving a payload-to-weight ratio of 4.6. 

TDMA enables the selection of higher-torque motors by reducing actuator count under the same weight constraint. The chosen servo provides 0.88$\,\mathrm{N\,m}$ rated torque, 2.26$\,\mathrm{N\,m}$ peak dynamic torque, and 4.90$\,\mathrm{N\,m}$ stall torque. Combined with the transmission design, this results in higher  actuation force and payload capacity compared to conventional mechanisms.

The workspace of the MuxArm were evaluated via Monte Carlo sampling ($10^9$ samples), constrained by the geometric limits of the backbone discs. Dexterity, defined as the orientation-coverage ratio within a $10^\circ$ tolerance, is projected onto the $XZ$ plane due to the manipulator's central symmetry (Fig.~\ref{fig:load}(c)). It peaks in the mid-radius region but diminishes near the workspace boundaries, such as in deeply folded or fully extended configurations.

\begin{figure}[t]
    \centering
    \includegraphics[width=0.9\linewidth]{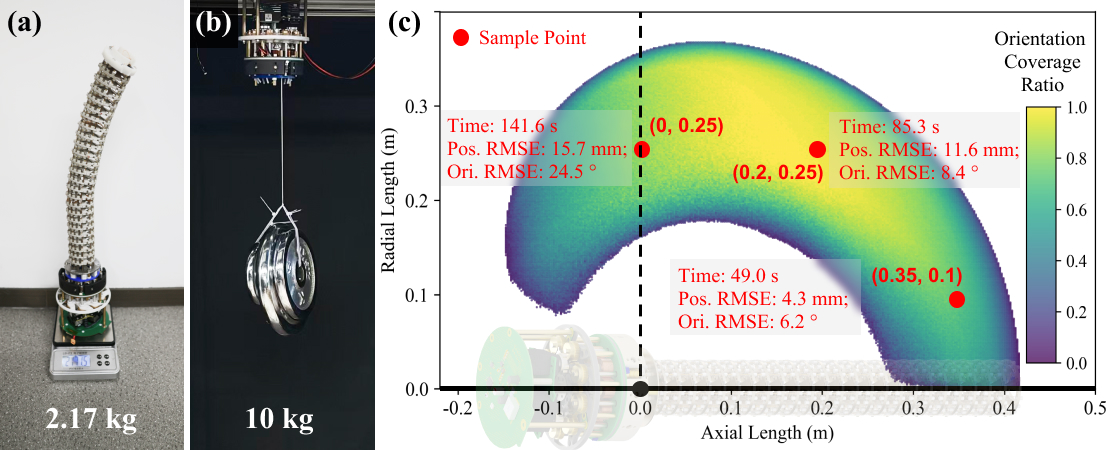}
    \caption{Performance evaluation of MuxArm.
(a) Self-weight measurement of 2.17 kg.
(b) Lifting test with a 10 kg payload.
(c) Workspace demonstration. The color map indicates the dexterity across the reachable workspace. Selected sample points illustrate the accuracy and time-to-reach performance.}
    \label{fig:load}
\end{figure}

For physical validation, MuxArm executed representative spatial patterns, including opposing bends, C-, and S-shapes (Fig.~\ref{fig:exp}(a)). All measured poses accurately fall within the predicted theoretical envelope. Quantitative repeatability evaluations (Fig.~\ref{fig:load}(c)) further demonstrate high accuracy in extended poses. The expected error increase at deeply folded boundaries stems from unmodeled friction and compliance under extreme bending, rather than TDMA switching limits. This confirms that the TDMA mechanism provides reliable, full-DOF authority and decoupled control for each section. We also open-sourced a Python-based simulator (\url{https://github.com/guo-cq/MuxArm}). It includes the dynamic visualization of configuration transitions and kinematics validation of MuxArm.

\subsection{Cost Analysis}

The system is low-cost, with a total cost of only \$541.85. Motors account for most of the cost. The TDMA reduces actuator cost by about 88\% compared with using individual high-torque compact servos, e.g., nine DYNAMIXEL XM540-W270-T units at \$482.89 each.

\begin{table}[h]
\centering
\caption{Cost of TDMA module (USD)}
\label{tab:cost}
    \renewcommand{\arraystretch}{1} 
\begin{tabular}{c c c l}
\hline
\textbf{Type} & \textbf{Item} & \textbf{~~Num~~} & \textbf{Price} \\
\hline
\multirow{2}{*}{Motors}
 & HTM4538 motor & 1 & \$109.60 \\
 & RX8U50HM servos & 3 & \$338.02~~~ \\
\hline
\multirow{9}{*}{\makecell{Standard\\Parts}}
 & Cylindrical gears & 18 & \$14.82 \\
 & Worm gears & 18 & \$13.58 \\
 & Releasing rods & 9 & \$2.64 \\
 & Copper columns & 14 & \$0.55 \\
 & Magnets & 9 & \$1.50 \\
 & Bearings & 9 & \$2.86 \\
 & Springs & 9 & \$1.50 \\
 & Screws &  & \$0.55 \\
\hline
\multirow{6}{*}{\makecell{~~Non-standard~~\\Parts}}
 & PCB boards & 5 & \$6.25 \\
 & ~~Enameled wire coils~~ & 9 & \$4.32 \\
 & Plastic-steel support & 1 & \$28.17 \\
 & 3D-printed parts &  & \$12.35 \\
 & PTFE tube &  & \$0.14 \\
 & Other materials &  & \$5.00 \\
\hline
\textbf{Total Cost}& & & \textbf{\$541.85} \\
\hline
\end{tabular}
\end{table}

\begin{figure*}[t]
    \centering
    \includegraphics[width=0.90\linewidth]{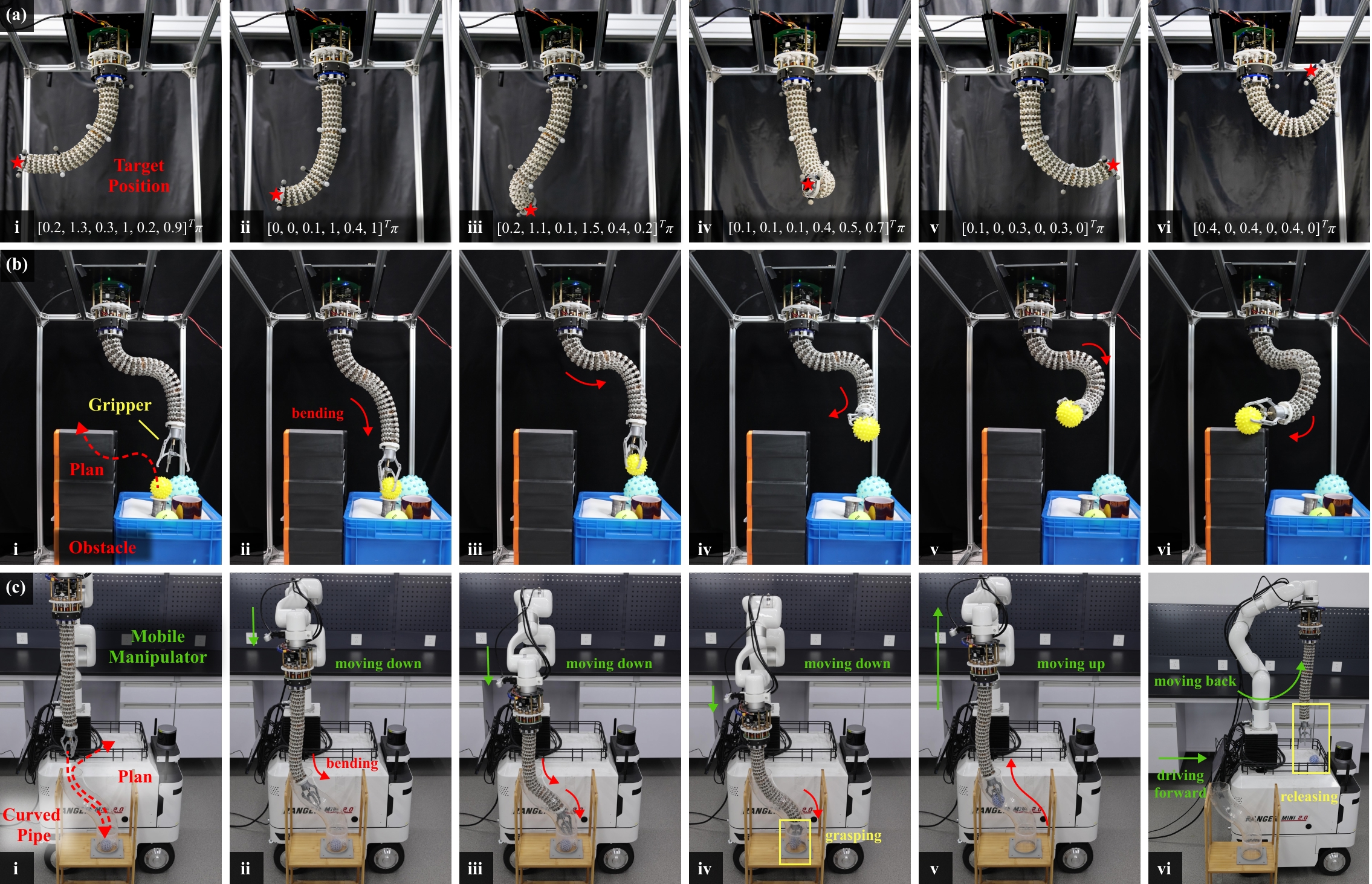}
    \caption{Experimental demonstrations validating the versatility of MuxArm.
(a) Workspace validation showing the large reachable end-effector space.
(b) Object grasping in a cluttered environment with obstacles.
(c) Grasping and extraction inside an S-shaped bent pipe using a mobile manipulator.}
    \label{fig:exp}
\end{figure*}

\subsection{Manipulation Tasks}

To assess the practicality of TDMA, we conduct two manipulation tasks in real environments. The tasks demonstrate redundancy-based obstacle avoidance in clutter, and operation in a confined pipe using a mobile manipulator.

\textbf{\textit{Task 1: cluttered grasping.}} MuxArm is fixed on a frame and equipped with a parallel gripper. A raised cabinet serves as the obstacle, and an adjacent box holds several objects. The goal is to grasp the yellow ball and place it above the obstacle. Fig.~\ref{fig:exp}(b) shows steps i–vi: the arm grasps the object, retracts by posture control, then rotates and places it on the target. In particular, steps iv–v show the extreme, tight S-shaped posture required for the task.

\textbf{\textit{Task 2: confined mobile manipulation.}} Leveraging the lightweight structure, MuxArm is mounted as the end effector of a mobile manipulator (AgileX Ranger Mini 2.0), as shown in Fig.~\ref{fig:exp}(c). A transparent S-shaped pipe lies on the floor with a small ball at the bottom. The mobile platform navigates with onboard lidar, and the manipulator aligns to the pipe mouth using an RGB–D camera (Intel D435i). The platform then holds position while the manipulator and MuxArm execute insertion (ii–iii), grasping (iv), extraction (v), releasing and departure (vi). A high-level planner uses the pipe shape for a coarse path, the manipulator provides gross motion, and MuxArm regulates local shape. For example, the sections bend sequentially while the arm first descends and then ascends to follow the local curvature in the insertion process.

Both tasks were completed successfully, demonstrating the system's capability to operate effectively in challenging environments. TDMA provides reliable multi-DOF control that exploits redundancy for obstacle avoidance and shape conformity in narrow passages. The manipulator actively adapts its configuration to conform to environmental constraints while executing planned trajectories.

\section{Conclusion}

This paper presents a practical approach for tendon-driven continuum robots that addresses the tradeoff between lightweight construction and fault-tolerant operation in aerospace applications. Implemented through a vertically-stacked rotational selection structure, TDMA enables four motors to control nine tendons across six degrees of freedom, achieving 55\% actuator reduction and 88\% cost reduction. The MuxArm manipulator demonstrates a self-weight of 2.17 kg, actuator driving capacity of 10 kg, and positioning accuracy up to 1\% of arm length. Statistical validation confirms the system maintains full kinematic accuracy under partial servo failures. TDMA represents a time-for-space tradeoff, as it introduces longer execution times but achieves mass efficiency and fault-tolerant ability, which is suitable for aerospace applications. 

Future work will advance TDMA toward practical deployment in three aspects. First, dynamic modeling research will establish comprehensive models accounting for cable elasticity, friction, and time-varying inertia to improve control accuracy under varying loads. Second, reliability enhancements including dual-axial-motor redundancy, fail-safe mechanisms, inline force sensors for real-time tension monitoring, and cable health monitoring will address aerospace safety requirements, while optimized scheduling algorithms will reduce task execution time. Third, field trials in aerospace environments and platform integration are essential for system certification. Beyond aerospace, further exploration will investigate its applications in other weight-constrained domains including aerial manipulation, mobile robotics, and hazardous environment operations.

\vspace{-3mm}

\bibliographystyle{IEEEtran}
\bibliography{ref}
\vspace{-16mm}

\begin{IEEEbiography}[{\includegraphics[width=1in,height=1.2in,clip,keepaspectratio]{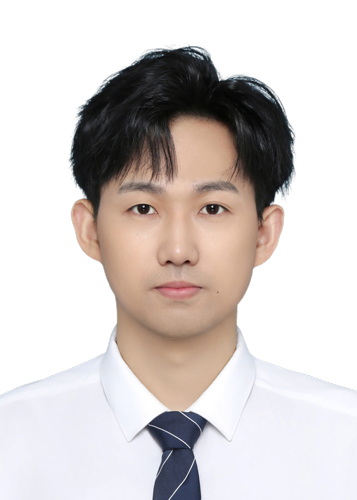}}]{Shoujie Li}
received the B.Eng. degree in electronic information engineering from the College of Oceanography and Space Informatics, China University of Petroleum,
Tsingtao, China, in 2020. He received Ph.D. degree from Tsinghua Shenzhen International Graduate School, Tsinghua University, Shenzhen, China, in 2025. He is working as a research fellow at Nanyang Technological University.

His research interests include tactile perception, grasping, and machine learning.
\end{IEEEbiography}
\vspace{-17mm}

\begin{IEEEbiography}[{\includegraphics[width=1in,height=1.2in,clip,keepaspectratio]{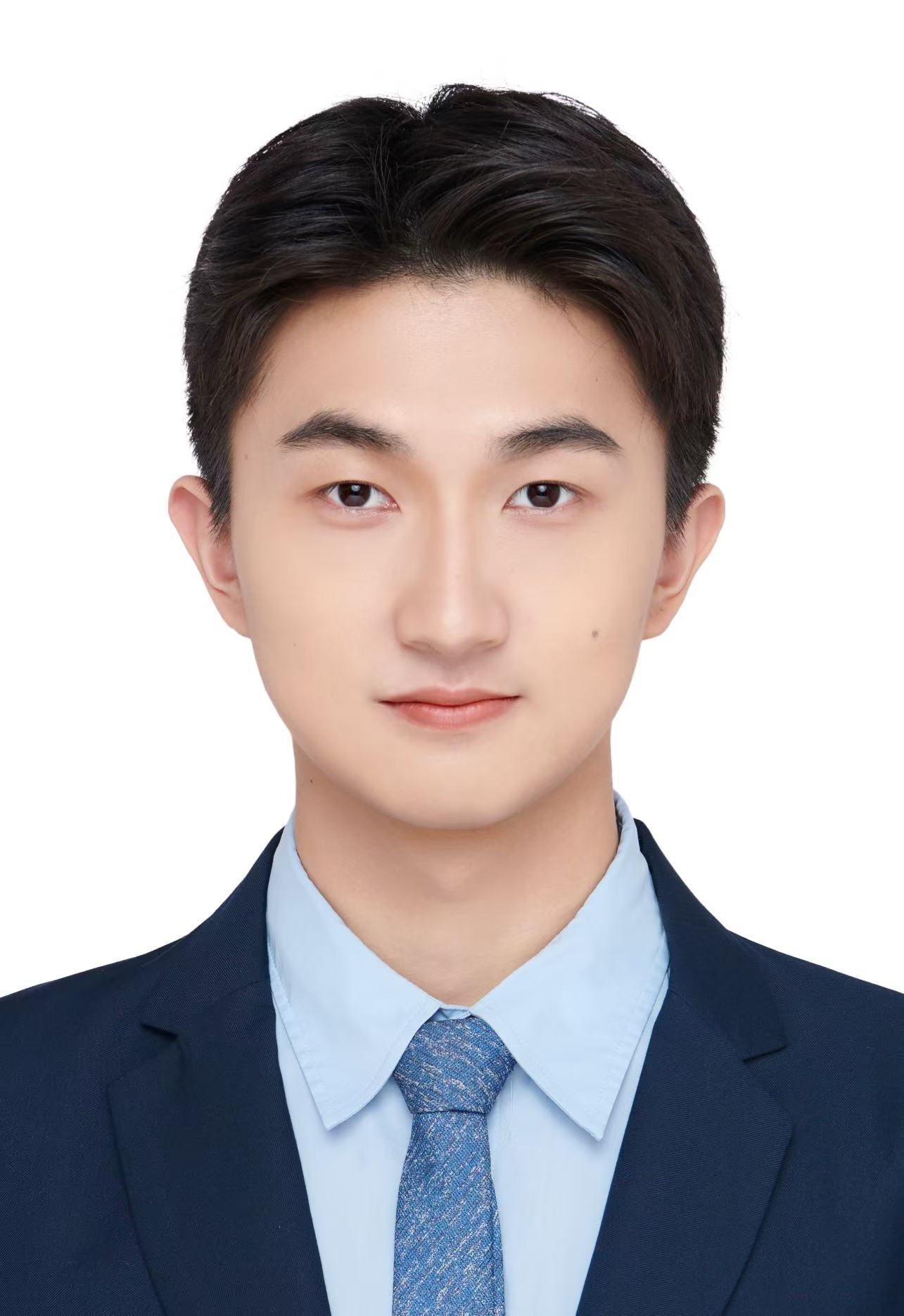}}]{Chanqing Guo} received the B.Eng. degree in Robotic Engineering from Shien-Ming Wu School of Intelligent Engineering, South China University of Technology, Guangzhou, China. He is currently pursuing the M.S. degree in Data Science and Information Technology at Tsinghua Shenzhen International Graduate School, Tsinghua University, Shenzhen, China.

His research interests include robotic manipulation and tactile perception.

\end{IEEEbiography}
\vspace{-18mm}

\begin{IEEEbiography}[{\includegraphics[width=1in,height=1.2in,clip,keepaspectratio]{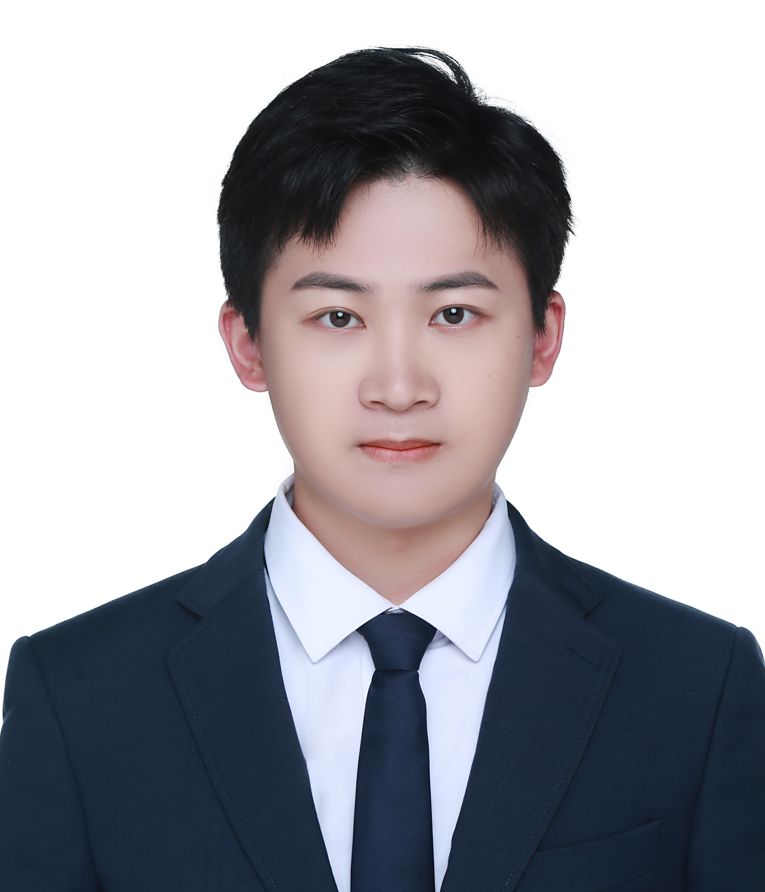}}]{Jianle Xu}
received the B.S. degree in Agricultural Mechanisation and Automation from Hainan University, Hainan, China, in 2023. He is currently working toward the M.S.degree in Tsinghua Shenzhen International Graduate School, Tsinghua University, Shenzhen, China.

His research interests include robot dexterous hands and electronic devices.
\end{IEEEbiography}
\vspace{-18mm}

\begin{IEEEbiography}[{\includegraphics[width=1in,height=1.2in,clip,keepaspectratio]{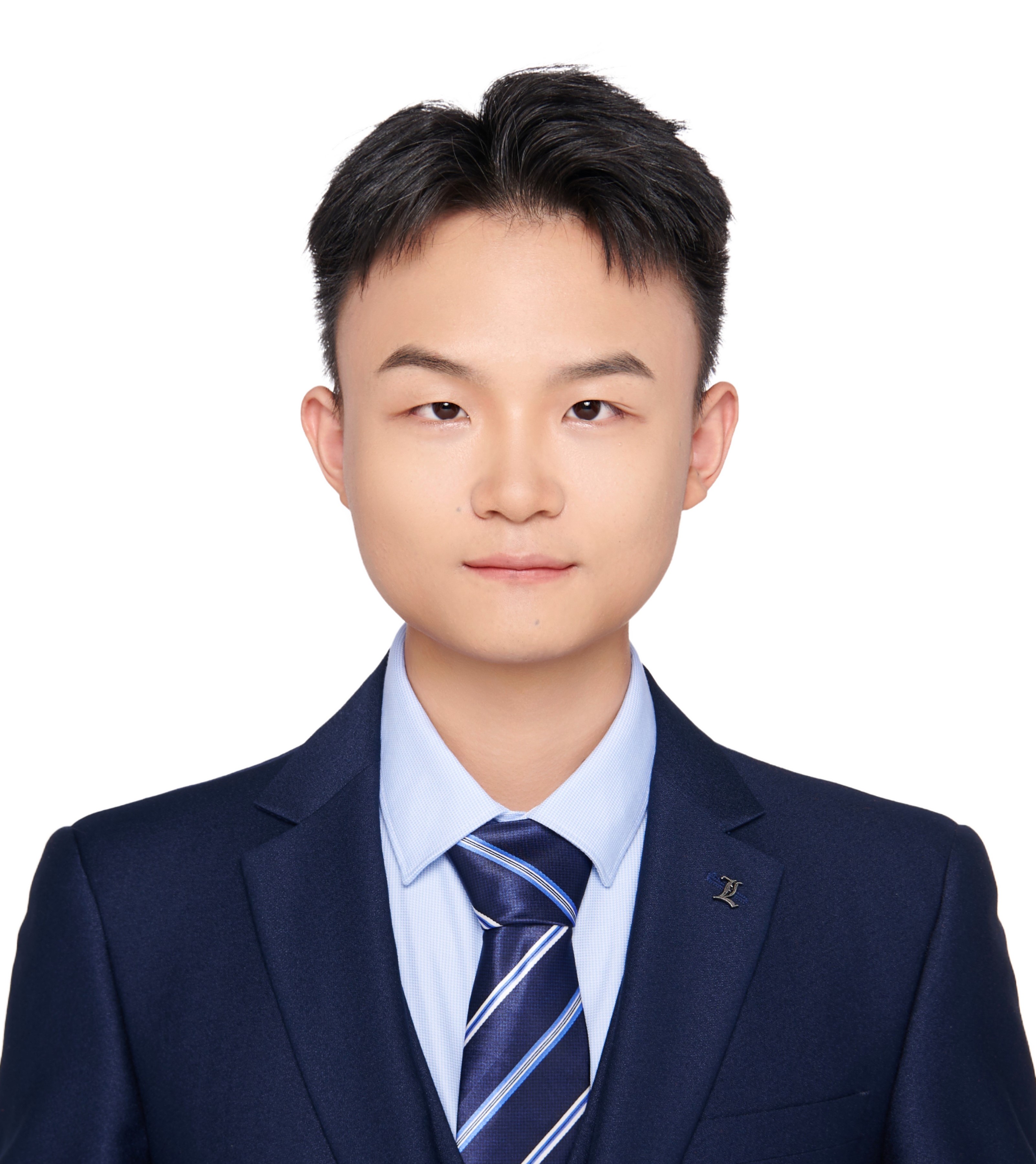}}]{Hong Luo}
 received the B.S. degree in Mechanical Design, Manufacturing, and Automation from Harbin Institute of Technology, Harbin, China. He is currently working toward the M.S. degree at Tsinghua Shenzhen International Graduate School, Tsinghua University, Shenzhen, China.

His research interests include the design and control methods of cable-driven robotic arms.
\end{IEEEbiography}
\vspace{-18mm}

\begin{IEEEbiography}
[{\includegraphics[width=1in,height=1.2in,clip,keepaspectratio]{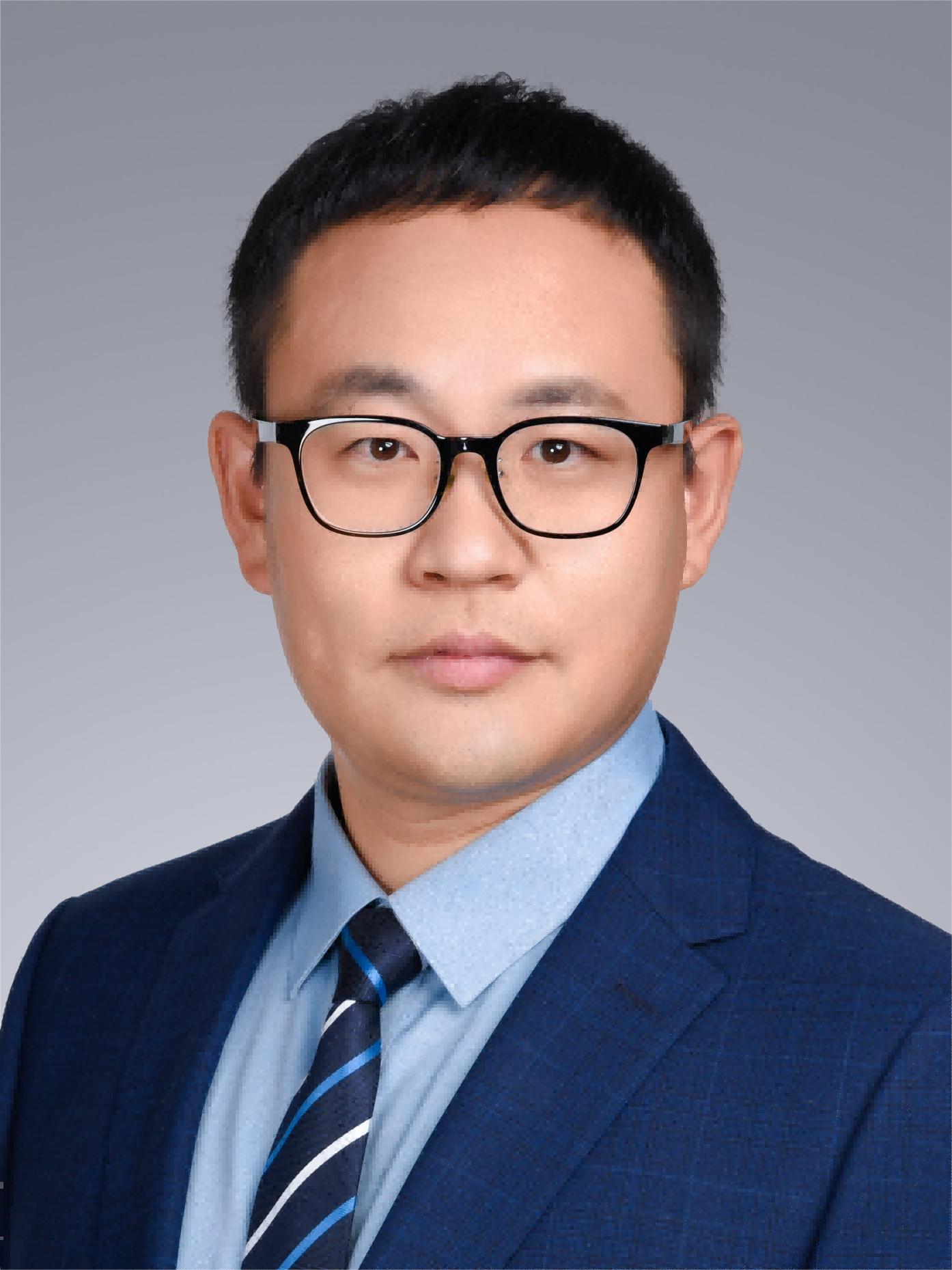}}]{Xueqian Wang}
received the B.E. degree in mechanical design, manufacturing and Automation, Harbin University of Science and Technology, Harbin, China, in 2003, the M.Sc. degree in mechatronic engineering and the  Ph.D. degree in control science and engineering from the Harbin Institute of Technology (HIT), Harbin, China, in 2005 and 2010, respectively. From June 2010 to February 2014, he was the Postdoc Research Fellow with the HIT. 

His research interests include dynamics modeling, control, and teleoperation of robotic systems.
 
\end{IEEEbiography}
\vspace{-18mm}

\begin{IEEEbiography}
[{\includegraphics[width=1in,height=1.2in,clip,keepaspectratio]{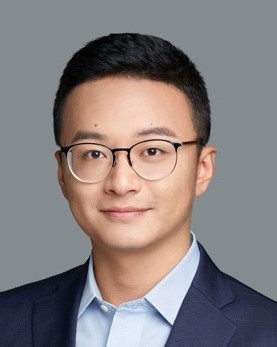}}]{Wenbo Ding}
 received the B.S. and Ph.D degrees (Hons.) from Tsinghua University in 2011 and 2016, respectively. He worked as a postdoctoral research fellow at Georgia Tech under the supervision of Professor Z. L. Wang from 2016 to 2019. He is now an associate professor and PhD supervisor at Tsinghua Shenzhen International Graduate School, Tsinghua University.
 
 His research interests include flexible electronics, tactile sensors, embodied intelligence and AI4S.
\end{IEEEbiography}
\vspace{-18mm}

\begin{IEEEbiography}
[{\includegraphics[width=1in,height=1.2in,clip,keepaspectratio]{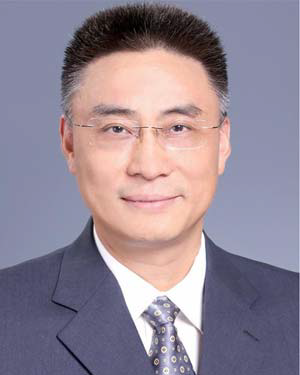}}]{Bin Liang}
 (Senior Member, IEEE) received the
Ph.D. degree in mechanical automation from the Department of Precision Instrument, Tsinghua University, Beijing, China, in 1994. Since 2007, he has been a Professor with the Department of Automation,Tsinghua University. 

His research interests include modeling and control of dynamic systems.

\end{IEEEbiography}
\vspace{-18mm}

\end{document}